\documentclass[sigconf]{acmart}

\AtBeginDocument{%
  }

\setcopyright{acmcopyright}
\copyrightyear{2023}
\acmYear{2023}
\acmDOI{XXXXXXX.XXXXXXX}

\acmConference[Conference acronym 'XX]{}{ }{ }

\usepackage{multirow}

\begin{document}

\title{Autotuning Apache TVM-based Scientific Applications Using Bayesian Optimization}


\author{Xingfu Wu} \affiliation{\institution{Argonne National Laboratory, Lemont, IL, USA}\country{}} \email{xingfu.wu@anl.gov}
\author{Praveen Paramasivam} \affiliation{\institution{University of South Dakota, Vermillion, SD, USA} \country{}}\email{praveen.paramasivam@coyotes.usd.edu}
\author{Valerie Taylor} \affiliation{\institution{Argonne National Laboratory, Lemont, IL, USA}\country{}} \email{vtaylor@anl.gov}

\renewcommand{\shortauthors}{Wu et al.}

\begin{abstract}
Apache TVM (Tensor Virtual Machine), an open source machine learning compiler framework designed to optimize computations across various hardware platforms, provides an opportunity to improve the performance of dense matrix factorizations such as LU (Lower–Upper) decomposition and Cholesky decomposition on GPUs and AI (Artificial Intelligence) accelerators. In this paper, we propose a new TVM autotuning framework using Bayesian Optimization and use the TVM tensor expression language to implement linear algebra kernels such as LU, Cholesky, and 3mm.  We use these scientific computation kernels to evaluate the effectiveness of our methods on a GPU cluster, called Swing, at Argonne National Laboratory. We compare the proposed autotuning framework with the TVM autotuning framework AutoTVM with four tuners and find that our framework outperforms AutoTVM in most cases.
  
\end{abstract}

\maketitle

\section{Introduction}

Current deep learning (DL) frameworks, such as Pytorch and TensorFlow, rely on a computational graph intermediate representation to implement optimizations. While graph-level representations are a good fit for high-level optimizations in DL, they are too high level to optimize tensor operators under a diverse set of hardware platforms such as CPUs, GPUs, and AI (Artificial Intelligence) accelerators. Generally speaking, deep learning workloads have high arithmetic intensity, which can typically be decomposed into tensor operators like matrix-matrix multiplication. These natural decompositions have led to the recent trend of adding tensor compute primitives \cite{Jouppi17, Chen18}. Apache TVM (Tensor Virtual Machine) \cite{TVM, Chen18} is an open source machine learning compiler framework designed to optimize computations across various hardware platforms. It introduces a tensor expression (TE) language to build operators, provides program transformation primitives that generate different versions of the program with various optimizations, supports an automated program optimization framework AutoTVM \cite{AutoTVM, Chen18b} to find optimized tensor operators, and provides a graph rewriter to take full advantage of high- and operator-level optimizations.

Dense matrix factorizations, such as LU and Cholesky, are widely used for scientific applications that require solving systems of linear equations, eigenvalues, and linear least squares problems. Such real-world scientific applications often take days, weeks, and even months to execute on tens and even hundreds of thousands of compute nodes with CPUs. TVM provides an opportunity to improve the performance of these dense matrix factorizations on GPUs and AI accelerators. In this paper, we propose a new autotuning framework using Bayesian Optimization in ytopt \cite{WU23, WU22} and use the TVM tensor expression language to implement linear algebra kernels such as LU, Cholesky, and 3mm from PolyBench 4.2 \cite{YP16}.  We use these scientific kernels to evaluate the effectiveness of our methods on a GPU cluster, called Swing \cite{Swing}, at Argonne National Laboratory.

In this paper, we make the following contributions:
\begin{itemize}
\item We propose a new autotuning framework for TVM-based scientific tensor applications using Bayesian Optimization.
\item We use TVM to implement scientific kernels such as LU, Cholesky, and 3mm.
\item We evaluate the effectiveness of the proposed autotuning framework and compare its performance with AutoTVM. 
\end{itemize}

The remainder of this paper is organized as follows. Section 2 describes the backgrounds about Apache TVM and ytopt. Section 3 proposes a new autotuning framework using ytopt. Section 4 discusses linear algebra benchmarks and their TVM TE implementations. Section 5 presents the experimental results and analyzes and compares the performance. Section 6 summarizes this paper and briefly discusses some future work.

\section{Backgrounds}
In this section, we briefly discuss some backgrounds about Apache TVM and ytopt.

\subsection{Apache TVM and its Tuning Framework}

\begin{figure}[ht]
  \centering
  \includegraphics[width=\linewidth]{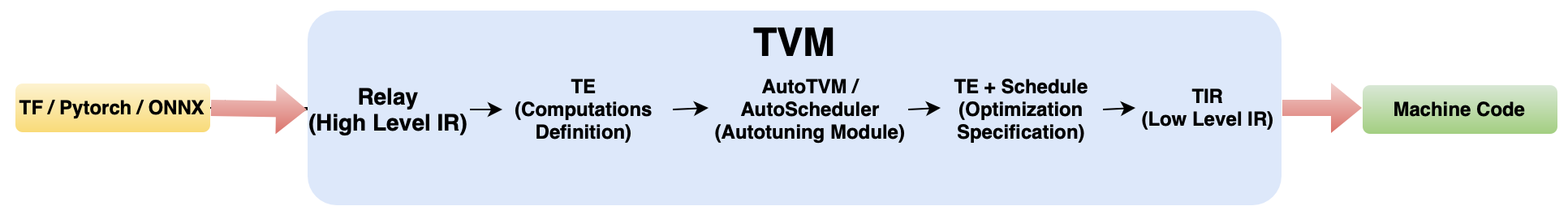}
  \caption{TVM Optimization Compiler and Tuning Framework}
  \label{fig:1}
\end{figure}

TVM \cite{TVM, Chen18} is an open source machine learning compiler framework designed to optimize computations across various hardware platforms such as CPUs, GPUs, and ML accelerators. Figure \ref{fig:1} shows the TVM optimization compiler framework. It supports models from popular deep learning frameworks such as TensorFlow, PyTorch, and ONNX, making it versatile and widely applicable. When a model is imported into Apache TVM, it undergoes conversion into a high-level intermediate representation using TVM’s high-level model language Relay \cite{Roesch19} which is a functional language and intermediate representation (IR) for neural networks. Relay applies graph-level optimization passes to optimize the model. TVM provides the Tensor Expression (TE) language which is a domain-specific language for describing tensor computations. After applying the high-level optimizations, Relay runs FuseOps pass to partition the model into many small subgraphs and lowers the subgraphs to the TE representation. TE provides several schedule primitives to specify low-level loop optimizations, such as loop titling, vectorization, parallelization, unrolling and fusion. A schedule specifies the low-level loop optimizations for an operator or subgraph defined in TE. 

For the auto-tuning module, Apache TVM leverages two approaches for auto-tuning to automate the process of optimizing tensor computation to search for the best schedule for loop optimizations, AutoTVM \cite{AutoTVM, Chen18b} and AutoScheduler \cite{AutoSD, Zheng20, Xing21}. 
AutoTVM relies on predefined tunable parameters search space to optimize the model, while AutoScheduler automatically generates the search space by analyzing the computation definition. The auto-tuning modules search for the best schedule and compare them with statistical cost models and on-device measurements. 

After the auto-tuning process, Apache TVM generates a JSON file containing all the schedules, from which the best schedule is selected based on the tuning results. Then each TE subgraph is transformed into Tensor Intermediate Representation (TIR) and further optimization through low-level optimization passes. The optimized TIR is eventually lowered to the target compiler of the hardware platform, resulting in a final optimized code ready for deployment in production. Ultimately, the compiler-specific generated code can be translated into machine code, ensuring the optimized model can be efficiently executed on the target hardware platform with a lightweight TVM runtime. 

In this paper, we investigate the effectiveness of the auto-tuning modules using AutoTVM and apply Bayesian Optimization to autotune tensor computation for TVM and compare their performance.

\subsection{ytopt: A ML-based Autotuning Tool Using Bayesian Optimization}

ytopt \cite{WU23, WU22} is a machine-learning-based search software package that consists of sampling a small number of input parameter configurations, evaluating them, and progressively fitting a surrogate model over the input-output space until exhausting the user-defined time or the maximum number of evaluations. The package is built based on Bayesian Optimization that solves optimization problems.

\begin{figure}[ht]
\center
 \includegraphics[width=\linewidth]{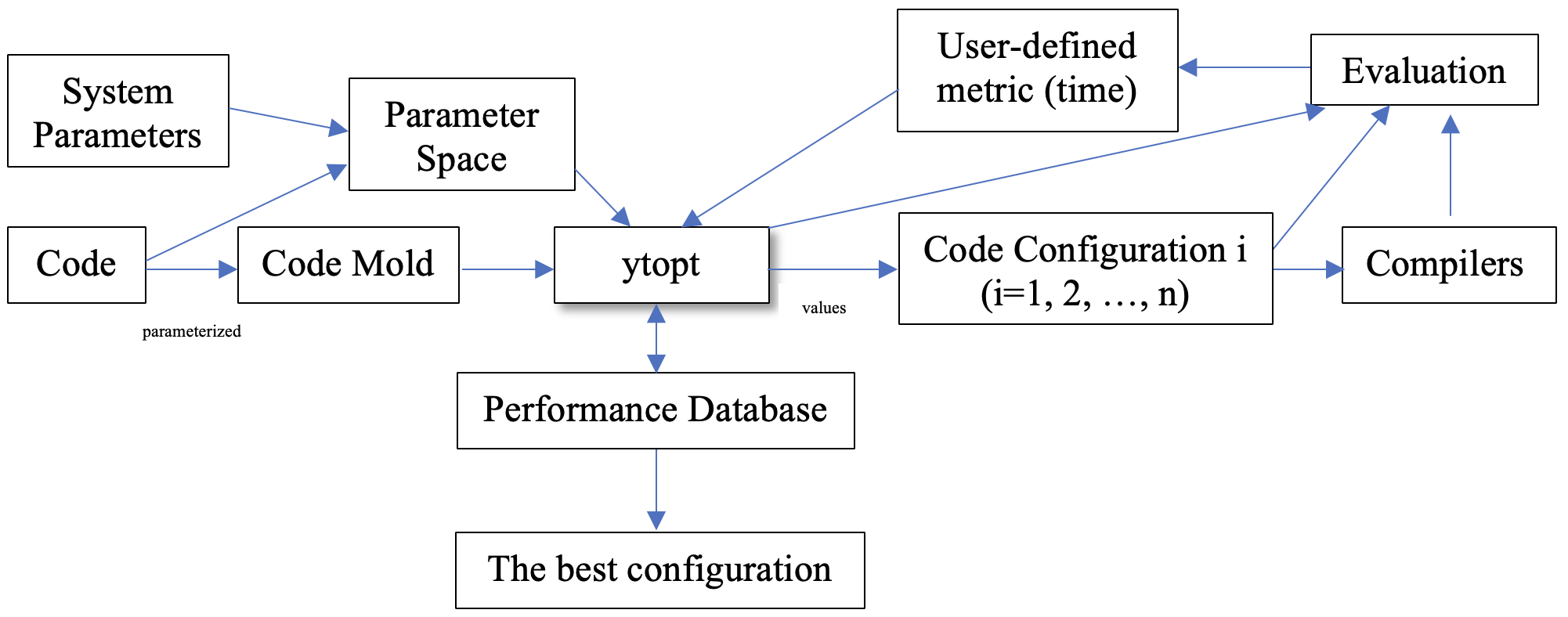}
 \caption{ytopt Autotuning Framework}
\label{fig:pf}
\end{figure}

Figure \ref{fig:pf} presents the framework for autotuning various applications. The application runtime is the primary user-defined metric. We analyze an application code to identify the important tunable application and system parameters to define the parameter space using ConfigSpace \cite{CFS}.
We use the tunable parameters to parameterize an application code as a code mold. 
ytopt starts with the user-defined parameter space, the code mold, and user-defined interface that specifies how to evaluate the code mold with a particular parameter configuration. 
The search method within ytopt uses Bayesian optimization, where a dynamically updated Random Forest surrogate model that learns the relationship between the configurations and the performance metric, is used to balance exploration and exploitation of the search space. In the exploration phase, the search evaluates parameter configurations that improve the quality of the surrogate model, and in the exploitation phase, the search evaluates parameter configurations that are closer to the previously found high-performing parameter configurations. The balance is achieved through the use of the lower confidence bound (LCB) acquisition function that uses the surrogate models' predicted values of the unevaluated parameter configurations and the corresponding uncertainty values.

\section{Proposed Autotuning Framework}

For the auto-tuning module, Apache TVM leverages two approaches for auto-tuning: AutoTVM and AutoScheduler in Figure \ref{fig:1}. AutoTVM relies on predefined tunable parameters search space to optimize the model, while AutoScheduler automatically generates the search space by analyzing the computation definition. Because AutoScheduler’s search space is not explicit, it is hard to find the search space to be used to compare with different tuning strategies. In this paper, we focus on AutoTVM because it requires predefined tunable parameter search space for four different tuner strategies which are as follows:
\begin{itemize}
\item RandomTuner: enumerate the space in a random order;
\item GridSearchTuner: enumerate the space in a grid search order; 
\item GATuner: use a genetic algorithm to search through  the space; 
\item XGBTuner: train a XGBoost model \cite{Chen16} to predict the runtime of lowered IR and pick the next batch according to the prediction. 
\end{itemize}

In our recent work \cite{WU22, WU23}, we developed and enhanced our autotuning framework ytopt to tune performance and energy for various scientific applications on large scale HPC systems. One question is, can we replace AutoTVM with ytopt to autotune the TVM-based scientific applications more efficiently? This is the motivation of this work. 

\begin{figure}[ht]
  \centering
  \includegraphics[width=\linewidth]{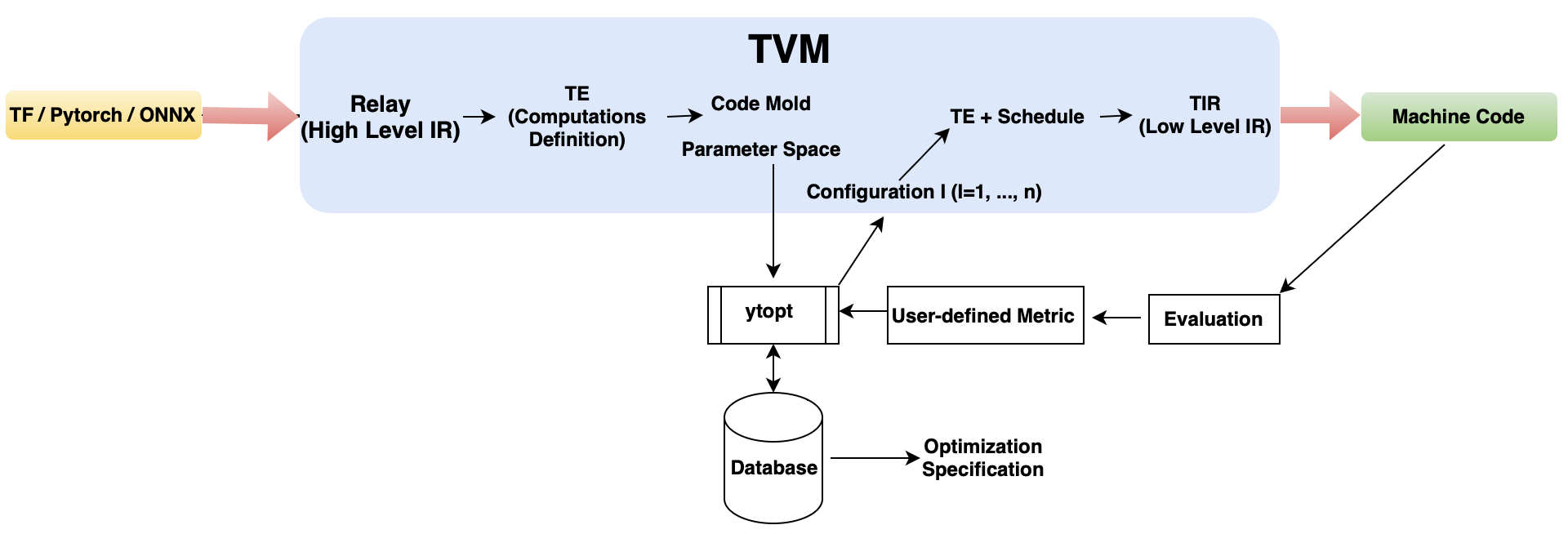}
  \caption{Proposed Autotuning Framework using Bayesian Optimization}
  \label{fig:2}
\end{figure}

Figure \ref{fig:2} presents the proposed TVM autotuning framework using ytopt. We basically replace the autotuning modules in Figure \ref{fig:1} with the ytopt module. Based on the TE code, we identify tunable parameters and use them to define the parameter space and to paramerize the TE code to generate its code mold. ytopt starts with the user-defined parameter space, the code mold, and user-defined interface that specifies how to evaluate the code mold with a particular parameter configuration. 

The iterative phase of the autotuning framework has the following steps: 
\begin{itemize}
\item [Step1]  Bayesian optimization in ytopt selects a parameter configuration for evaluation. 
\item [Step2] The code mold is configured with the selected configuration to generate a new TE code. 
\item [Step3] The new code is compiled with other codes needed to generate an executable (machine code).
\item [Step4]  The machine code is executed to evaluate the application with the selected parameter configuration
\item [Step5] The resulting application runtime (user-defined metric) is sent back to ytopt and recorded in the performance database. 
\end{itemize}

Steps 1--5 are repeated until the maximum number of evaluations $n$ or the wall-clock time is exhausted for the autotuning run. In the end, we query the performance database to output the optimization specification for the best configuration. 

In this way, the proposed autotuning framework identifies the best configuration for the TE code on a target system. This is different from AutoTVM with four tuners, where AutoTVM identifies the best configuration and passes it to the TE code to generate the machine code for evaluation on the target system.

\section{Linear Algebra Benchmarks}

PolyBench 4.2 \cite{YP16} is a benchmark suite of 30 numerical computations with static control flow, extracted from operations in various application domains (linear algebra computations, image processing, physics simulation, and data mining). For the sake of simplicity, in this paper, we focus on linear algebra kernel 3mm  and linear algebra solvers Cholesky and LU. 


Cholesky is Cholesky decomposition in Linear Algebra Solvers, which decomposes a matrix to triangular matrices and entails A = $L*L^T$, where L: NxN lower triangular matrix, and A: N × N positive-definite matrix . We use the following large datasets: large dataset (N 2000) and extralarge dataset (N 4000)  for our case study.

LU is LU (Lower–Upper) decomposition without pivoting in linear algebra solvers and entails A = L*U, where L is an NxN lower triangular matrix and U is an NxN upper triangular matrix. We use the following large datasets : large dataset (N 2000) and extralarge dataset (N 4000) for our case study.

3mm is one of the linear algebra kernels that consists of three matrix multiplications and entails G=(A*B)*(C*D), where A is a NxL matrix; B is a LxM matrix; C is an MxO matrix; and D is an OxP matrix. We use the following large datasets: large dataset (N 800, L 900, M 1000, O 1100, P 1200) and extralarge dataset (N 1600, L 1800, M 2000, O 2200, P 2400) for our case study.

We use Apache TVM to implement 3mm, Cholesky and LU based on the algorithms from the C implementation of PolyBench 4.2 and use them as the baselines to conduct the autotuning experiments. Table \ref{tab:ps} shows the parameter space size for each application. We use TVM TE (tensor expression) to implement these kernels. For simplicity, we mainly focus on a split optimization using a block size of the reordering. 

\begin{table}[ht]
\center
\caption{Parameter space for each application}
\begin{tabular}{|r|c|c|c|}
\hline
Kernels & Problem Size & Parameter Space    \\ 
\hline
\multirow{2}{*}{3mm} &  large &   74,649,600 \\ 
\cline{2-3} 
&  extralarge & 228,614,400   \\ 
\hline
\multirow{2}{*}{Cholesky} & large  & 400   \\ 
\cline{2-3}
& extralarge  & 576   \\
\hline
\multirow{2}{*}{LU} & large  & 400   \\ 
\cline{2-3}
& extralarge  & 576   \\
\hline
\end{tabular}
\label{tab:ps}
\end{table}

For instance, the basic TE implementation 3mm\_basic() of 3mm in Python is as follows:

{\scriptsize
\begin{verbatim}
def 3mm_basic(N, L, M, O, P, dtype):
    A = te.placeholder((N, L), name="A", dtype=dtype)
    B = te.placeholder((L, M), name="B", dtype=dtype)
    C = te.placeholder((M, O), name="C", dtype=dtype)
    D = te.placeholder((O, P), name="D", dtype=dtype)
    k = te.reduce_axis((0, L), name="k")
    l = te.reduce_axis((0, O), name="l")
    m = te.reduce_axis((0, M), name="m")
    E = te.compute((N, M), lambda i, j: te.sum(A[i, k] * B[k, j], axis=k), name="E")
    F = te.compute((M, P), lambda i, j: te.sum(C[i, l] * D[l, j], axis=l), name="F")
    G = te.compute((N, P), lambda i, j: te.sum(E[i, m] * F[m, j], axis=m), name="G")
    s1 = te.create_schedule(E.op)
    s2 = te.create_schedule(F.op)
    s3 = te.create_schedule(G.op)
    y, x = s1[E].op.axis
    k = s1[E].op.reduce_axis[0]
    y1, x1 = s2[F].op.axis
    l = s2[F].op.reduce_axis[0]
    y2, x2 = s3[G].op.axis
    m = s3[G].op.reduce_axis[0]
\end{verbatim}
{\color{red}

    yo, yi = s1[E].split(y, 8)
    
    xo, xi = s1[E].split(x, 8)
    
    yo1, yi1 = s2[F].split(y1, 8)
    
    xo1, xi1 = s2[F].split(x1, 8)
    
    yo2, yi2 = s3[G].split(y2, 8)
    
    xo2, xi2 = s3[G].split(x2, 8)
    }
\begin{verbatim}
    s1[E].reorder(yo, xo, k, yi, xi)
    s2[F].reorder(yo1, xo1, l, yi1, xi1)
    s3[G].reorder(yo2, xo2, m, yi2, xi2)
    return s3, [A, B, C, D, G]
\end{verbatim}
 }  

Where we use 8 as the loop tiling factor for both x and y, x1 and y1, and x2 and y2 for the statements with the red color, however, the best tiling factor depends on real hardware environment and input shape, therefore, we use the six tiling factors as tunable parameters to parameterize the code to generate a code mold with the only the following six different statements to replace the six statements with the red color in the 3mm\_basic() above: 
{\scriptsize
\begin{verbatim}
    yo, yi = s1[E].split(y, #P0)
    xo, xi = s1[E].split(x, #P1)
    yo1, yi1 = s2[F].split(y1, #P2)
    xo1, xi1 = s2[F].split(x1, #P3)
    yo2, yi2 = s3[G].split(y2, #P4)
    xo2, xi2 = s3[G].split(x2, #P5)
\end{verbatim}
 }  

We use Configspace to define the parameter space for 3mm with the extralarge problem size for ytopt as follows:
{\scriptsize
\begin{verbatim}
P0= CSH.OrdinalHyperparameter(name='P0', sequence= [1, 2, 4, 5, 8, 10, 16, 20, 25, 40, 50,
80, 100, 125, 200, 250, 400, 500, 1000, 2000])
P1= CSH.OrdinalHyperparameter(name='P1', sequence= [1, 2, 4, 5, 8, 10, 16, 20, 25, 32, 40,
50, 64, 80, 100, 160, 200, 320, 400, 800, 1600])
P2= CSH.OrdinalHyperparameter(name='P2', sequence= [1, 2, 3, 4, 5, 6, 8, 10, 12, 15, 16, 
20, 24, 25, 30, 32, 40, 48, 50, 60, 75, 80, 96, 100, 120, 150, 160, 200, 240, 300, 400, 
480, 600, 800, 1200, 2400])
P3= CSH.OrdinalHyperparameter(name='P3', sequence= [1, 2, 4, 5, 8, 10, 16, 20, 25, 40, 50,
80, 100, 125, 200, 250, 400, 500, 1000, 2000])
P4= CSH.OrdinalHyperparameter(name='P4', sequence= [1, 2, 3, 4, 5, 6, 8, 10, 12, 15, 16, 
20, 24, 25, 30, 32, 40, 48, 50, 60, 75, 80, 96, 100, 120, 150, 160, 200, 240, 300, 400, 
480, 600, 800, 1200, 2400])
P5= CSH.OrdinalHyperparameter(name='P5', sequence= [1, 2, 4, 5, 8, 10, 16, 20, 25, 32, 40,
50, 64, 80, 100, 160, 200, 320, 400, 800, 1600])
cs.add_hyperparameters([P0, P1, P2, P3, P4, P5])
\end{verbatim}
} 

Where the matrix sizes are 1600x2000 for matrix AxB, 2000x24000 for matrix CxD, and 1600x2400 for matrix ExF. We use the common factors of each matrix rank to define a set of candidate values for each tunable parameter to form a parameter space.

For AutoTVM, the parameter space is defined using the following statements to replace the six statements with the red color in the 3mm\_basic():

{\scriptsize
\begin{verbatim}
    cfg = autotvm.get_config()
    cfg.define_knob("tile_y", [1, 2, 4, 5, 8, 10, 16, 20, 25, 40, 50, 80, 100, 125, 
    200, 250, 400, 500, 1000, 2000])
    cfg.define_knob("tile_x", [1, 2, 4, 5, 8, 10, 16, 20, 25, 32, 40, 50, 64, 80, 100,
    160, 200, 320, 400, 800, 1600])
    cfg.define_knob("tile_y1", [1, 2, 3, 4, 5, 6, 8, 10, 12, 15, 16, 20, 24, 25, 30, 32, 40, 
    48, 50, 60, 75, 80, 96, 100, 120, 150, 160, 200, 240, 300, 400, 480, 600, 800, 1200, 2400])
    cfg.define_knob("tile_x1", [1, 2, 4, 5, 8, 10, 16, 20, 25, 40, 50, 80, 100, 125, 200,
    250, 400, 500, 1000, 2000])
    cfg.define_knob("tile_y2", [1, 2, 3, 4, 5, 6, 8, 10, 12, 15, 16, 20, 24, 25, 30, 32, 40, 
    48, 50, 60, 75, 80, 96, 100, 120, 150, 160, 200, 240, 300, 400, 480, 600, 800, 1200, 2400])
    cfg.define_knob("tile_x2", [1, 2, 4, 5, 8, 10, 16, 20, 25, 32, 40, 50, 64, 80, 100, 
    160, 200, 320, 400, 800, 1600])  
    yo, yi = s1[E].split(y, cfg["tile_y"].val)
    xo, xi = s1[E].split(x, cfg["tile_x"].val)
    yo1, yi1 = s2[F].split(y1, cfg["tile_y1"].val)
    xo1, xi1 = s2[F].split(x1, cfg["tile_x1"].val)
    yo2, yi2 = s3[G].split(y2, cfg["tile_y2"].val)
    xo2, xi2 = s3[G].split(x2, cfg["tile_x2"].val)
\end{verbatim}
 }  

Similarly, we use TVM to implement Cholesky and LU and their autotuning codes and define their parameter spaces.

\section{Performance Analysis}
In this section, we use TVM-based cholesky, lu and 3mm to conduct the autotuning experiments using AutoTVM with four tuners and ytopt on Argonne GPU cluster Swing \cite{Swing} and analyze and compare their performance. Swing has 2x AMD EPYC 7742 64-Core Processor and 8x NVIDIA A100 GPUS per node, and 1TB DDR memory per node and 40 GB HBM memory per GPU. For all autotuning experiments, we set just 100 evaluations then compare their best cases and the overall autotuning process time which is the total time spending in finishing 100 evaluations.

\begin{figure}[ht]
  \centering
  \includegraphics[width=\linewidth]{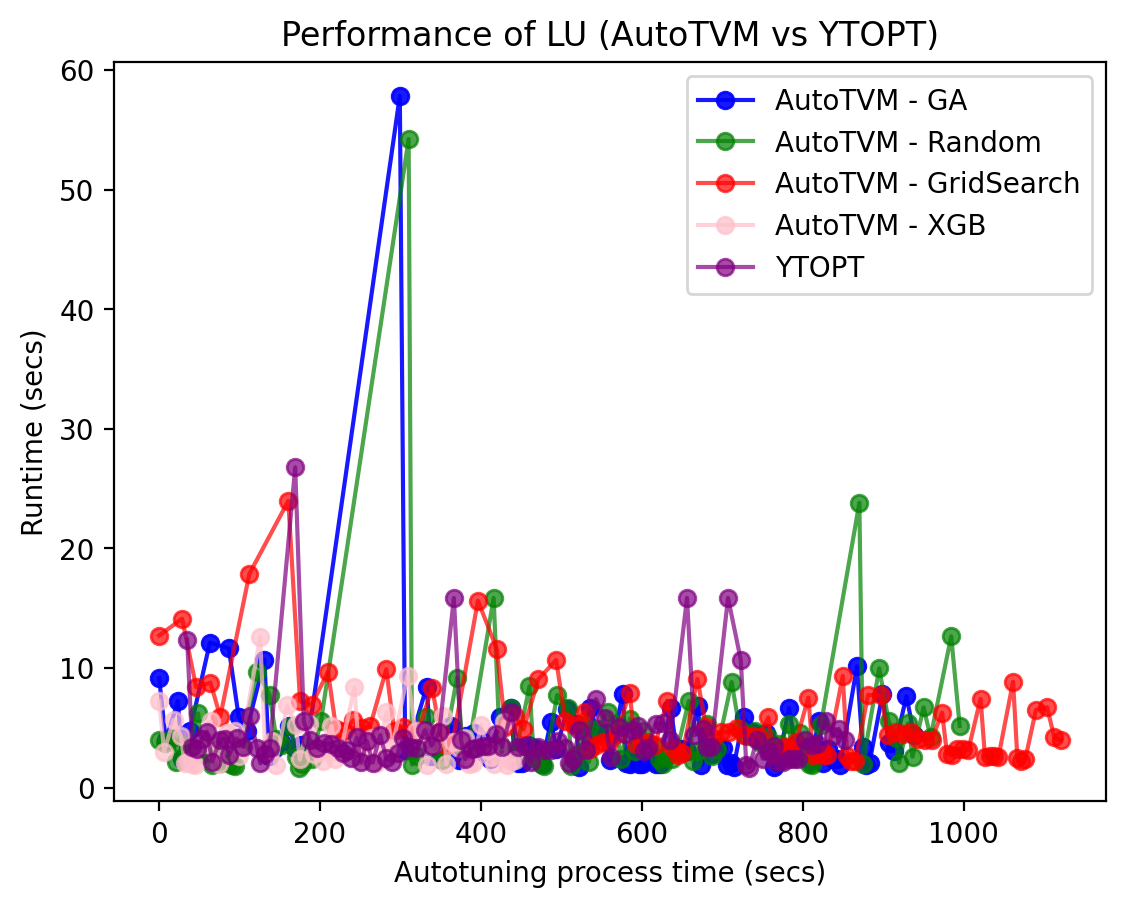}
  \caption{Performance comparison for LU with large problem size}
  \label{fig:7}
\end{figure}

\begin{figure}[ht]
  \centering
  \includegraphics[width=\linewidth]{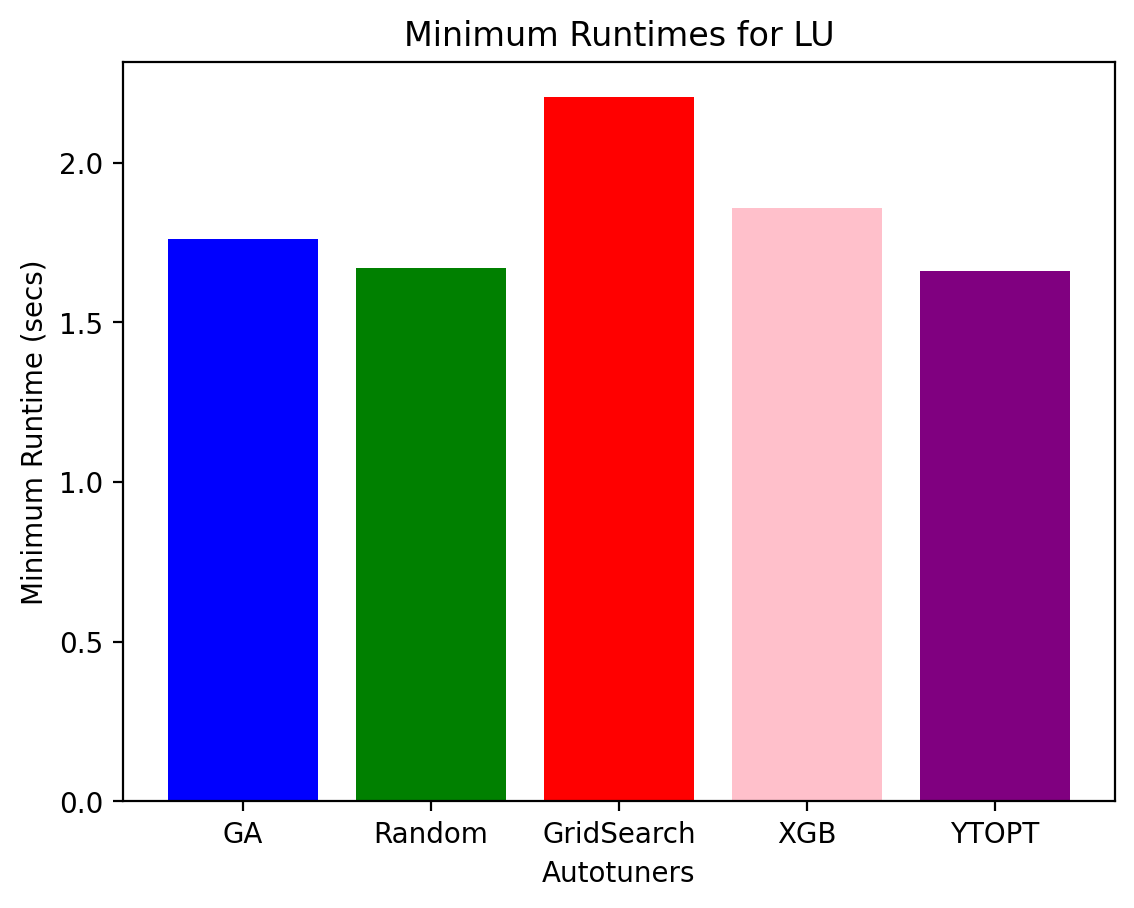}
   \setlength{\belowcaptionskip}{-16pt}
  \caption{Minimum runtimes for LU with large problem size}
  \label{fig:8}
\end{figure}

For LU, Figure \ref{fig:7} shows the autotuning process over time for LU with the large problem size using 4 AutoTVM tuners and ytopt. Each point represents one evaluation with its runtime (on y-axis). AutoTVM-GA stands for AutoTVM with GATuner (GA); AutoTVM-Random statnds for AutoTVM with RandomTuner (Random); AutoTVM-GridSearch stands for AutoTVM with GridSearchTuner (GridSearch); and AutoTVM-XGB stands for AutoTVM with XGBTuner (XGB). We observe that ytopt took the smallest time to finish 100 evaluations. Notice that XGBoot search tuner could only do at most 56 evaluations no matter how many evaluations are set for some reason. Figure \ref{fig:8} shows that ytopt outperforms 4 AutoTVM tuners to identify the best tensor size 400x50 to result in the smallest runtime (1.659s). 

\begin{figure}[ht]
  \centering
  \includegraphics[width=\linewidth]{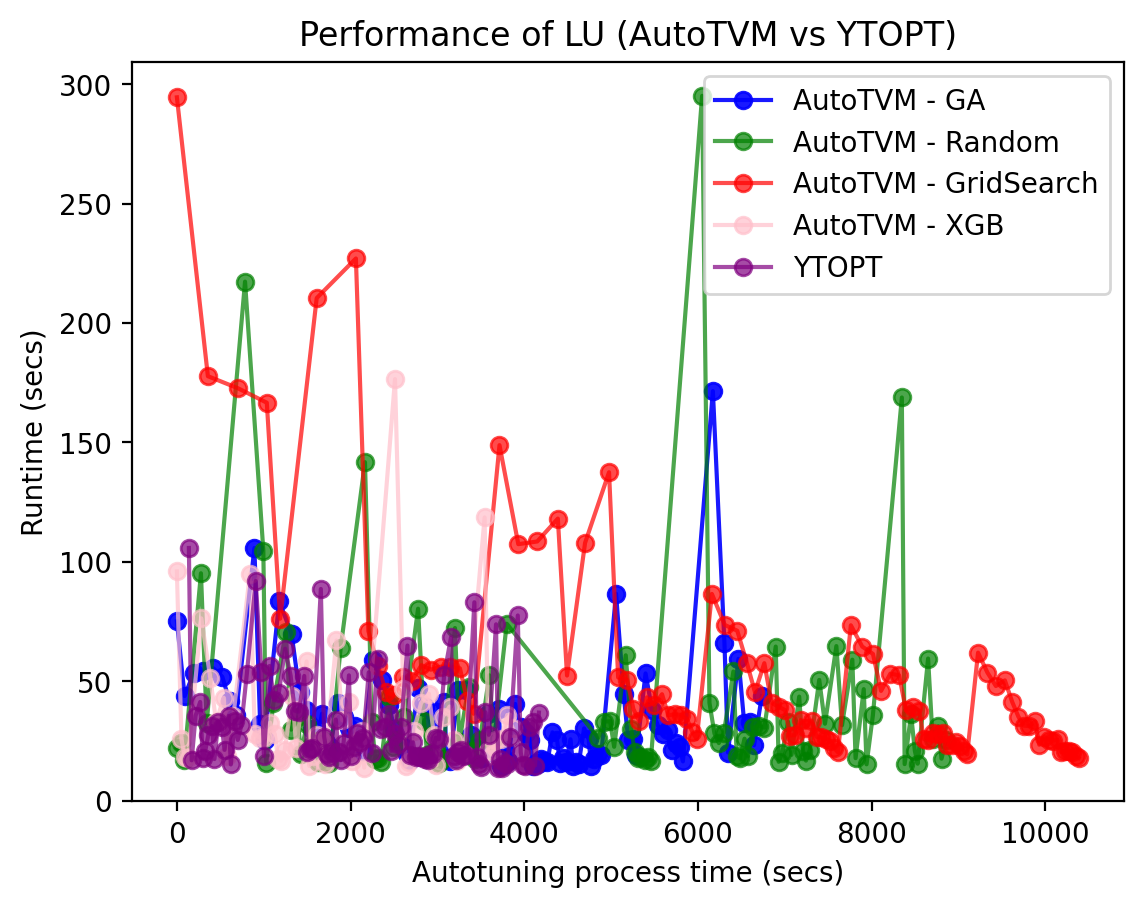}
  \caption{Performance comparison for LU with extralarge problem size}
  \label{fig:9}
\end{figure}

\begin{figure}[ht]
  \centering
  \includegraphics[width=\linewidth]{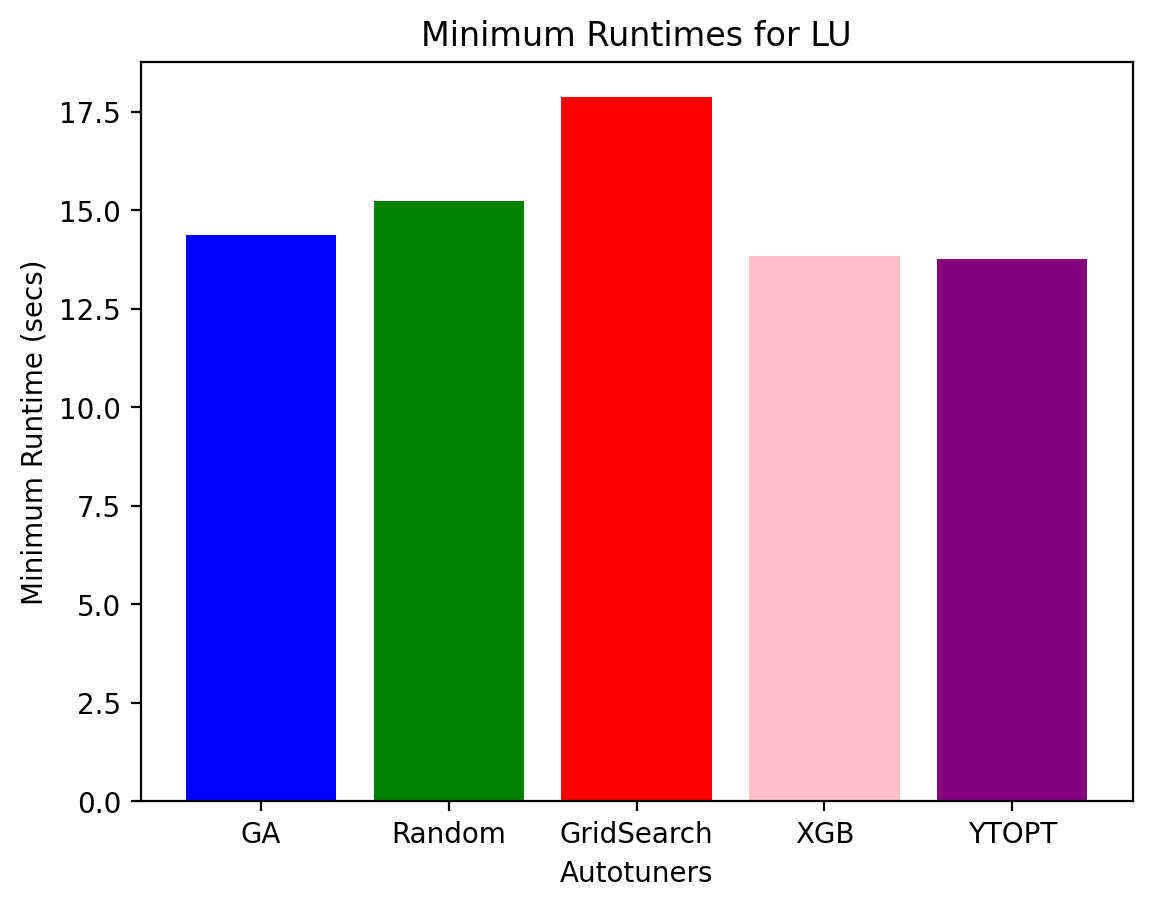}
  \caption{Minimum runtimes for LU with extralarge problem size}
  \label{fig:10}
\end{figure}

Similarly, for LU with the extralarge problem size, ytopt also outperformed 4 AutoTVM tuners shown in Figure \ref{fig:9} in the smallest autotuning process time. Figure \ref{fig:10} shows that ytopt identified the best tensor size 40x32 to result in the smallest runtime (13.77s). It is interesting to see the two different tensor sizes were identified, however, they both were not included in the same performance database because of the different parameter spaces.

\begin{figure}[ht]
  \centering
  \includegraphics[width=\linewidth]{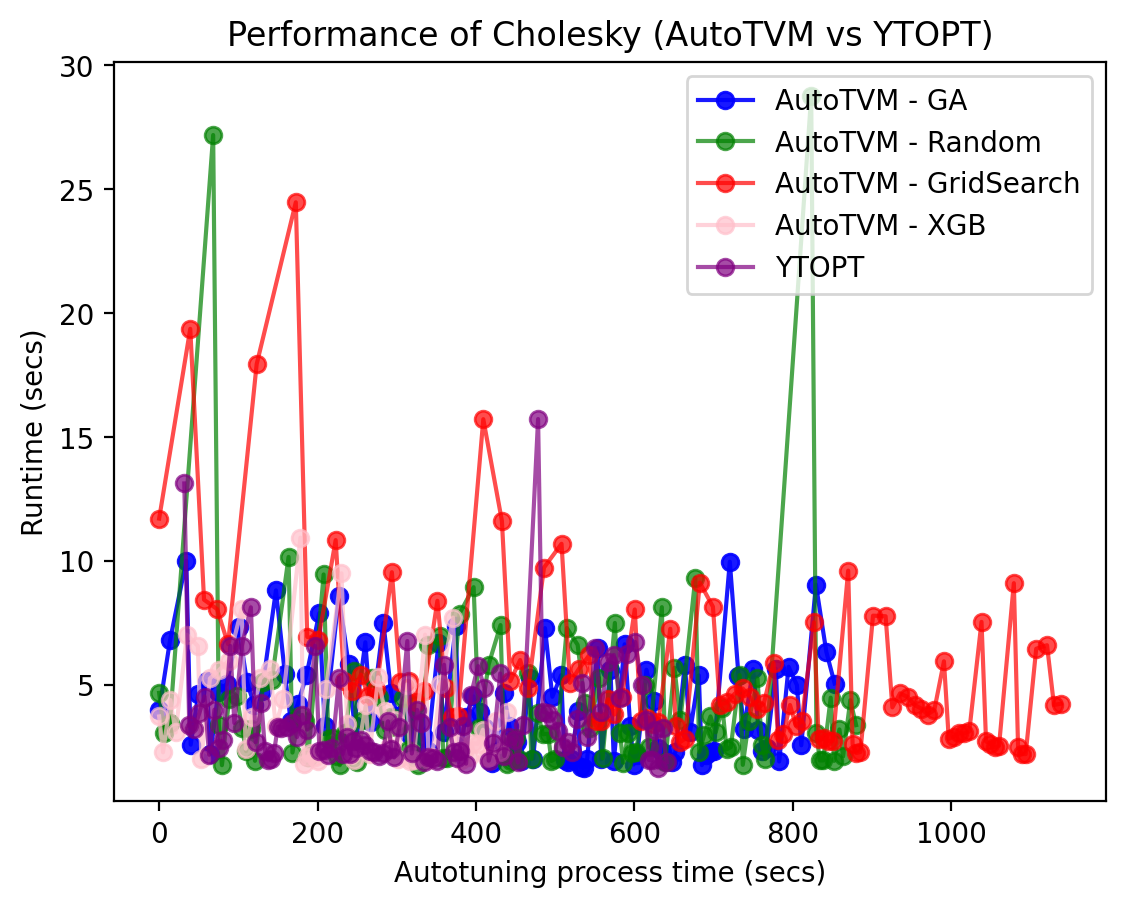}
  \caption{Performance comparison for Cholesky with large problem size}
  \label{fig:3}
\end{figure}

\begin{figure}[ht]
  \centering
  \includegraphics[width=\linewidth]{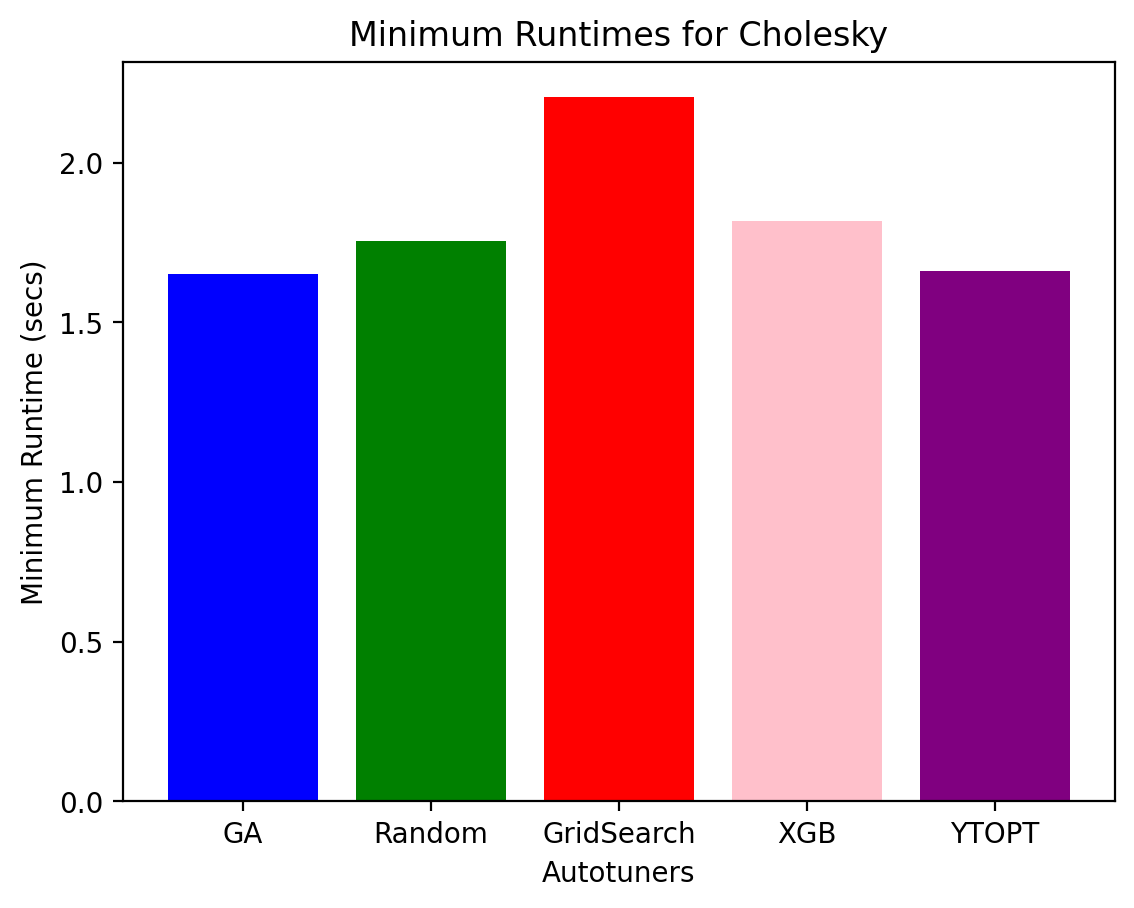}
  \setlength{\belowcaptionskip}{-16pt}
  \caption{Minimum runtimes for Cholesky with large problem size}
  \label{fig:4}
\end{figure}

For Cholesky, Figure \ref{fig:3} shows the autotuning process over time for cholesky with the large problem size using 4 AutoTVM tuners and ytopt. AutoTVM-GA results in the smallest runtime (1.65s) with the tensor size 50x50, but ytopt took much less time to finish 100 evaluations than AutoTVM-GA. Figure \ref{fig:4} shows that ytopt outperforms the other 3 AutoTVM tuners to identify the best tensor size 125x50 to result in the smallest runtime (1.66s). 

\begin{figure}[ht]
  \centering
  \includegraphics[width=\linewidth]{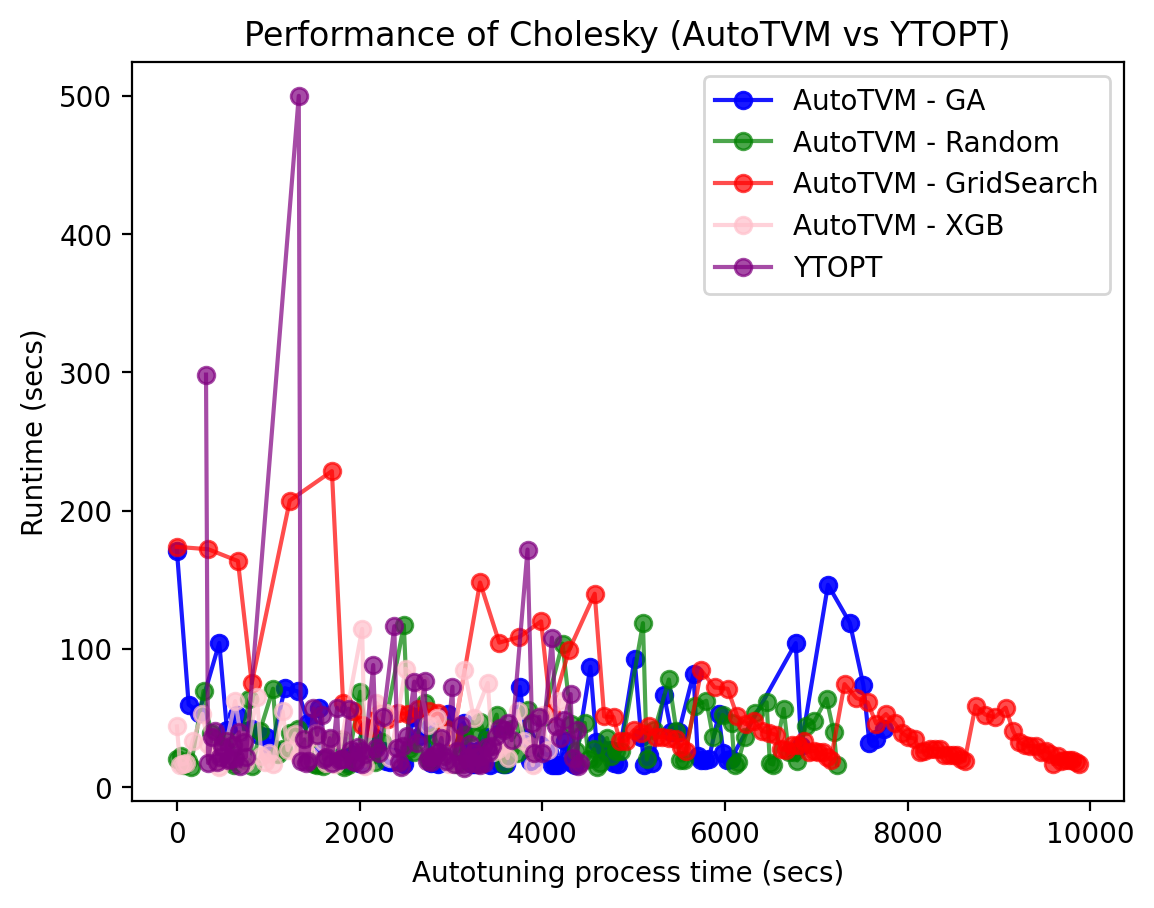}
  \caption{Performance comparison for Cholesky with extralarge problem size}
  \label{fig:5}
\end{figure}

\begin{figure}[ht]
  \centering
  \includegraphics[width=\linewidth]{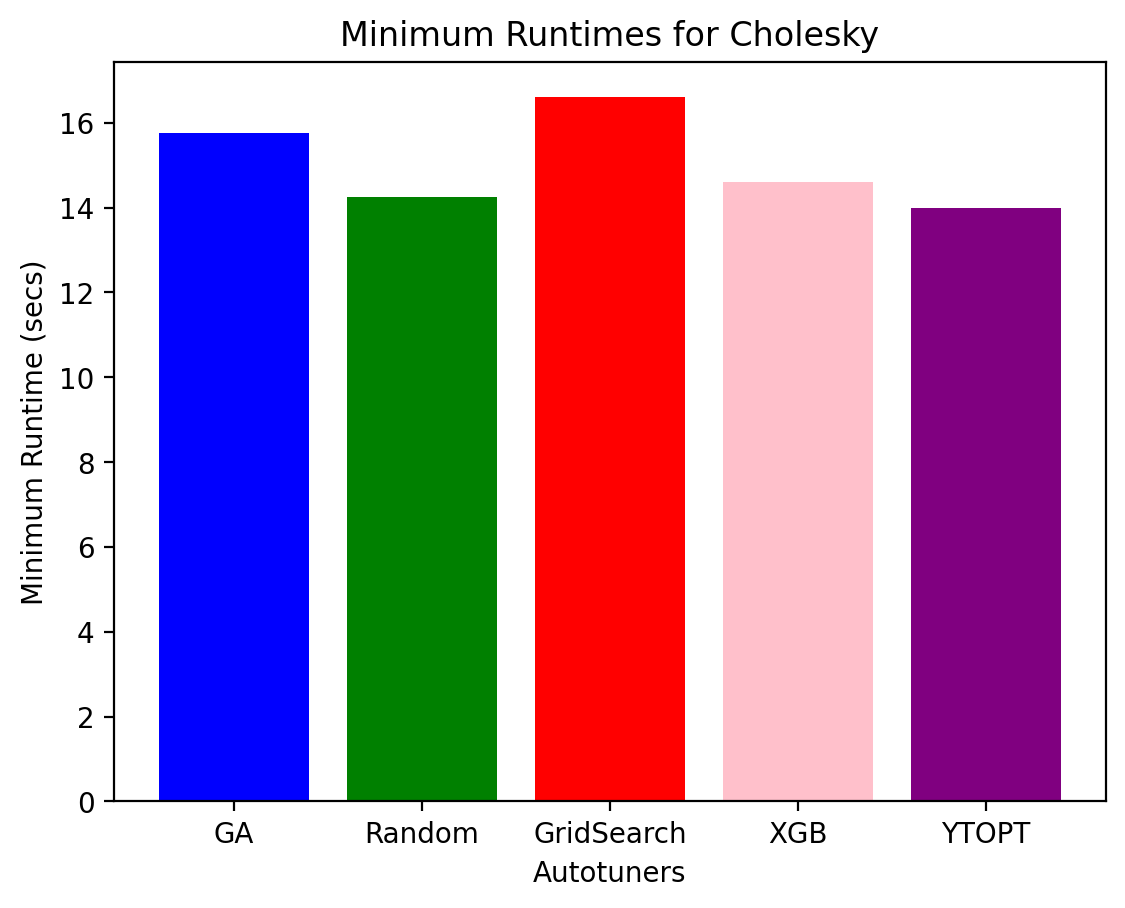}
  \caption{Minimum runtimes for Cholesky with extralarge problem size}
  \label{fig:6}
\end{figure}

Similarly, for Cholesky with the extralarge problem size, ytopt outperformed 4 AutoTVM tuners shown in Figure \ref{fig:5} in the smallest autotuning process time. Figure \ref{fig:6} shows that ytopt identified the best tensor size 80x32 to result in the smallest runtime (13.99s).  

\begin{figure}[ht]
  \centering
  \includegraphics[width=\linewidth]{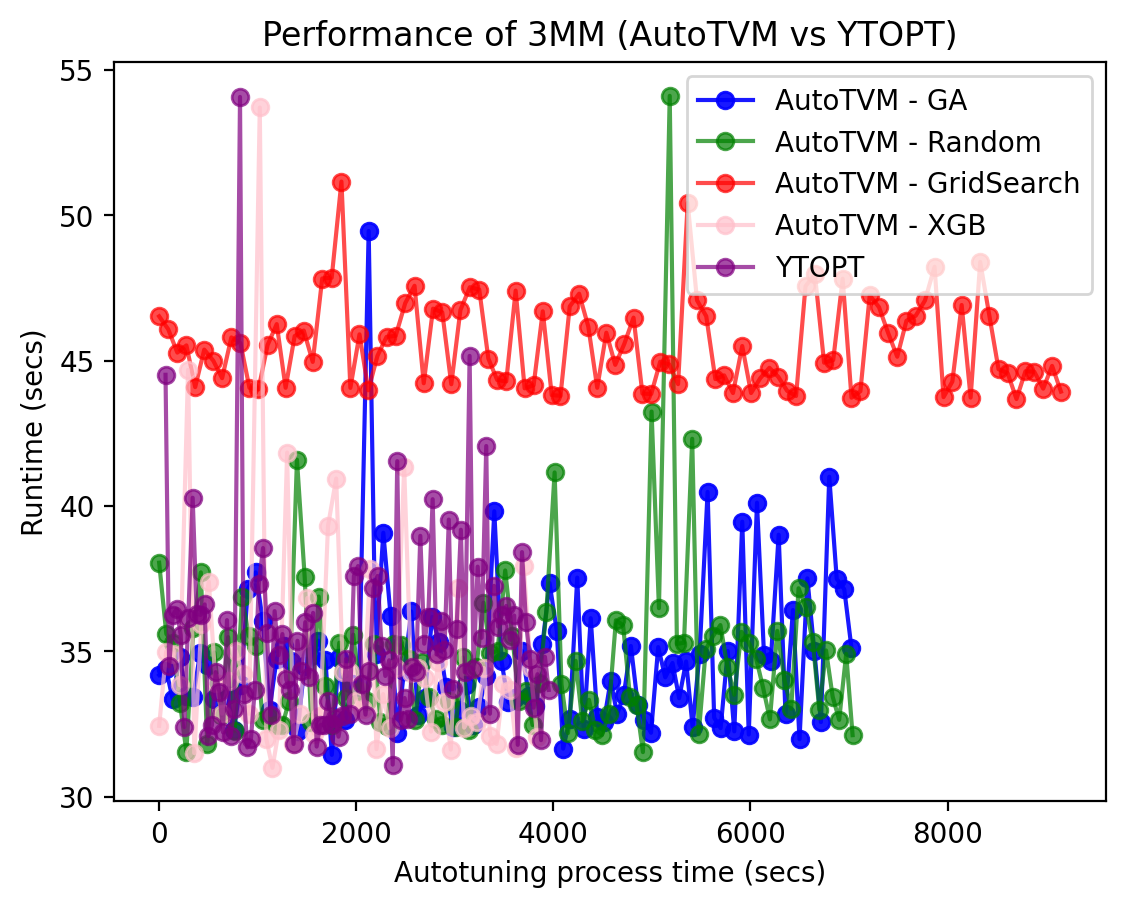}
  \caption{Performance comparison for 3mm with extralarge problem size}
  \label{fig:13}
\end{figure}

\begin{figure}[ht]
  \centering
  \includegraphics[width=\linewidth]{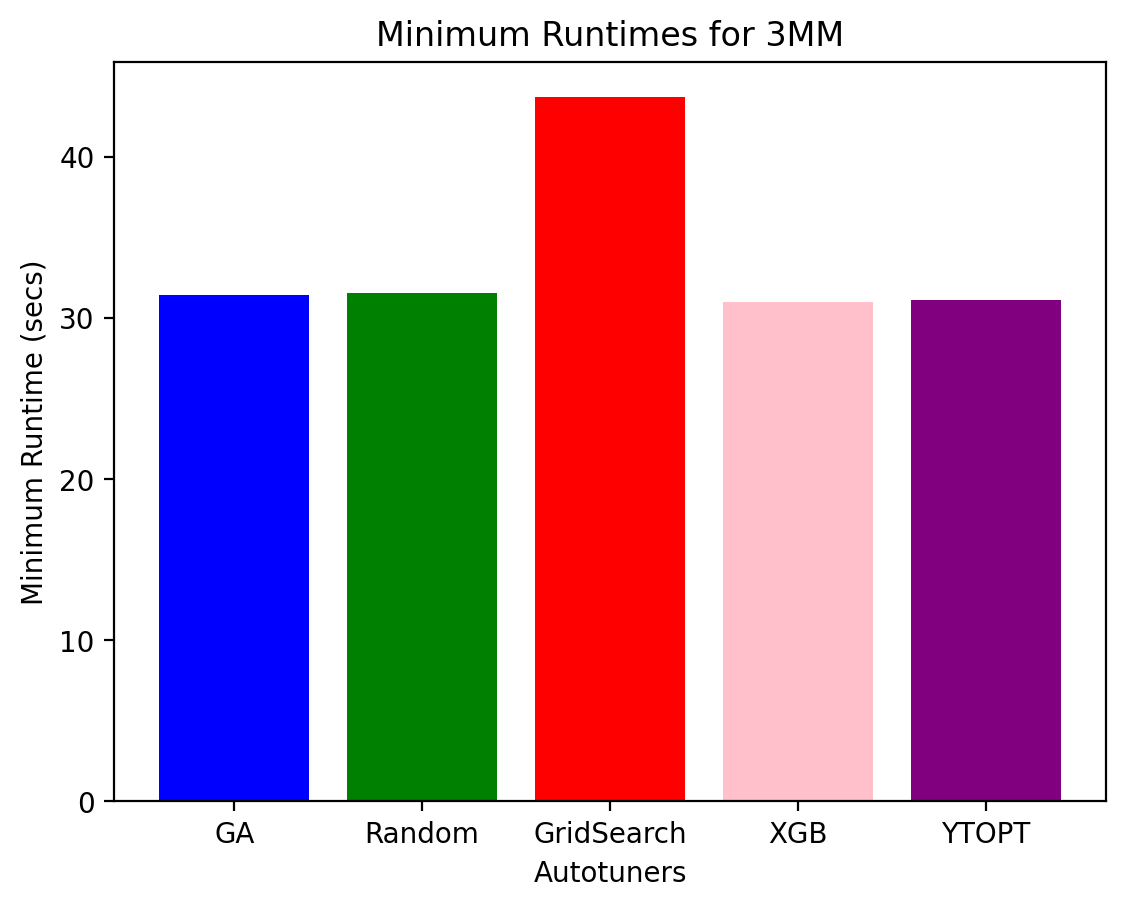}
  \caption{Minimum runtimes for 3mm with extralarge problem size}
  \label{fig:14}
\end{figure}

For 3mm, Figure \ref{fig:13} and Figure \ref{fig:14} show the autotuning process over time for 3mm with the extralarge problem size using 4 AutoTVM tuners and ytopt. AutoTVM-XGB results in the smallest runtime of 30.99s with the tensor size (1000x32, 600x2, 15x40), however, ytopt outperforms the other 3 AutoTVM tuners to identify the tensor sizes (1x5,120x25, 60x100) to result in the runtime of 31.1s.

Overall, for the effectiveness of AutoTVM, grid search tuner performed the worst for all the experiments; XGBoot search tuner could only do at most 56 evaluations no matter how many evaluations are set for some reason. ytopt outperformed AutoTVM in most cases and took the smallest autotuning process time with the extralarge problem sizes even though AutoTVM tuners use the statistical cost models to predict the next tiling factor. For the large problem sizes, because of using the statistical cost models to predict the next tiling factor AutoTVM takes relatively smaller autotuning process time for some cases.

\section{Conclusions}
TVM provides an opportunity to us to improve the performance of the dense matrix factorizations such as LU and Cholesky on GPUs and other accelerators. In this paper, we proposed a new TVM autotuning framework using Bayesian Optimization in ytopt, used TVM tensor expression language to implement linear algebra kernels such as LU, Cholesky, and 3mm, and then used these kernels to evaluate its effectiveness. We compared the proposed framework with the TVM autotuning framework AutoTVM with four tuners: GATuner, RandomTuner, GridSearchTuner and XGBTuner, and find that our framework outperformed AutoTVM in most cases.
For the effectiveness of AutoTVM with four tuners, grid search tuner performed the worst for all the experiments. The proposed autotuning framework outperformed AutoTVM and took the smallest autotuning process time in most cases. Future work will focus on using the proposed autotuning framework to tune deep learning models and operators using ResNet, MobileNet, and Deep Convolutional Generative Adversarial Networks on GPUs and AI accelerators.

\section{Acknowledgments}

This work was supported in part by DOE ECP PROTEAS-TUNE and in part by DOE ASCR RAPIDS2. We acknowledge the Argonne Laboratory Computing Resource Center (LCRC) for use of the GPU cluster Swing under LCRC project EE-ECP, and thank Prasanna Balaprakash from Oak Ridge National Laboratory for an initial discussion. This material is based upon work supported by the U.S. Department of Energy, Office of Science, under contract number DE-AC02-06CH11357. 
\newpage
\bibliographystyle{ACM-Reference-Format}
\bibliography{main}

\end{document}